%% file: bfpp.tex
\documentclass[twocolumn, switch]{article} 

\usepackage{preprint}

\usepackage{amsmath, amsthm, amssymb, amsfonts}

\usepackage[numbers,square]{natbib}
\usepackage[utf8]{inputenc}	
\usepackage[T1]{fontenc}	
\usepackage{xcolor}		
\usepackage[colorlinks = true,
            linkcolor = purple,
            urlcolor  = blue,
            citecolor = cyan,
            anchorcolor = black]{hyperref}	
\usepackage{booktabs} 		
\usepackage{nicefrac}		
\usepackage{microtype}		
\usepackage{lineno}		
\usepackage{float}			

\usepackage{lipsum}		

\usepackage{makecell}
\usepackage{ifthen}
\usepackage{fancyvrb}
\usepackage{multirow}
\usepackage{newfloat}
\DeclareFloatingEnvironment[name={Supplementary Figure}]{suppfigure}
\usepackage{sidecap}
\sidecaptionvpos{figure}{c}

\usepackage{titlesec}
\titlespacing\section{0pt}{12pt plus 3pt minus 3pt}{1pt plus 1pt minus 1pt}
\titlespacing\subsection{0pt}{10pt plus 3pt minus 3pt}{1pt plus 1pt minus 1pt}
\titlespacing\subsubsection{0pt}{8pt plus 3pt minus 3pt}{1pt plus 1pt minus 1pt}

\usepackage{subfig}

\title{BF++: a language for general-purpose program synthesis}

\usepackage{xwatermark}
\newwatermark[firstpage,color=gray!90,angle=0,scale=0.28, xpos=0in,ypos=-5in]{*correspondence: \texttt{v.liventsev@tue.nl}}

\usepackage{authblk}

\author[1\thanks{\tt{v.liventsev@tue.nl}}]{Vadim Liventsev}
\author[2]{Aki H{\"a}rm{\"a}}
\author[3]{Milan Petkovi{\'c}}
\affil[1,3]{Eindhoven University of Technology}
\affil[1,2,3]{Philips Research Eindhoven}

\begin{document}

\twocolumn[ 
  \begin{@twocolumnfalse} 
  
\maketitle

\begin{abstract}
Most state of the art decision systems based on Reinforcement Learning (RL) are data-driven black-box neural models, where it is often difficult to incorporate expert knowledge into the models or let experts review and validate the learned decision mechanisms. Knowledge-insertion and model review are important requirements in many applications involving human health and safety. One way to bridge the gap between data and knowledge driven systems is program synthesis: replacing a neural network that outputs decisions with a symbolic program generated by a neural network or by means of genetic programming. 
We propose a new programming language, BF++, designed specifically for automatic programming of agents in a Partially Observable Markov Decision Process (POMDP) setting and apply neural program synthesis to solve standard OpenAI Gym benchmarks. 
Source code is available at \url{https://github.com/vadim0x60/cibi}
\end{abstract}
\vspace{0.35cm}

  \end{@twocolumnfalse} 
] 



\input{1-intro}
\input{2-background}
\input{3-relatedwork}

\input{4-language}
\input{5-experiments}
\input{6-results}

\footnotesize
\section*{Acknowledgements}

This work was funded by the European Union’s Horizon 2020 research and innovation programme under grant agreement n° 812882. This work is part of "Personal Health Interfaces Leveraging HUman-MAchine Natural interactionS" (PhilHumans) project: \url{https://www.philhumans.eu}

\normalsize
\bibliography{references}


\end{document}

%% file: 1-intro.tex
\section{Introduction}
\label{sec:intro}

Reinforcement Learning (RL) has been applied successfully in fields like Energy, Finance and Robotics \cite{rl-applications}.
However, traditional approaches to Reinforcement Learning involve black box models that preclude any exchange of knowledge between experts and ML algorithms.
In safety-critical fields\footnote{Safety requirements in healthcare are the main motivation for our research. However, in this paper we use conventional OpenAI Gym benchmarks to enable comparison between methods} like Healthcare \cite{healthcare-rl} the ability to understand the decision algorithms induced by artificial intelligence, as well as to initialize the system using expert knowledge for an acceptable baseline performance, is required for acceptability.

In this work we focus on an alternative approach for RL based on program induction, known as Programmatically Interpretable Reinforcement Learning \cite{pirl}. 
We introduce \textbf{BF++}, a new programming language tailor-made for this approach (section \ref{sec:bf}). 
We then demonstrate that neural program synthesis with \textbf{BF++} can solve arbitrary reinforcement learning challenges and gives us an avenue for knowledge sharing between domain experts and data-driven models via the mechanism of \emph{expert inspiration} (section \ref{sec:methodology}).

%% file: 2-background.tex
\section{Background}
In this paper we define a Reinforcement Learning environment as Partially Observable Markov Decision Process \cite{pomdp1,pomdp2}: when at step $i$ the agent takes action $a_i \in A$ it has an impact on  the state of the environment $s_i \in S$ via distribution $p_s(s_{i+1} | s_i, a_i)$ of conditional probabilities of possible subsequent states. 
State is a latent variable that the agent cannot observe.
Instead, the agent can see an observation $o_i \in O$ which is a random variable that depends on the latent state via distribution $p_o(o_i | s_i, a_i)$.
$A$, $S$ and $O$ are sets of all possible actions, states and observations respectively.
Finally, at every step the agent observes a reward $r_i=R(s_i,a_i)$

Given this limited toolset, without full (or any) prior knowledge of how the agent's actions influence the the environment (distributions $p_s(s_{i+1} | s_i, a_i)$ and $p_o(o_i | s_i, a_i)$), the agent has to come up with a strategy that will maximize $n$-step return $R_n=\sum_{t=i}^{n} r_t$ where $n$ is the agent's planning horizon. It is, in the general sense, a hyperparameter, however if an environment has a limit on how many steps an episode can last, it is reasonable to set $n$ equal to the step limit. 

Conventional solutions \cite{thebook} introduce a parametrized \emph{policy function} $\pi_\phi(a|s)$ that defines agent's behavior as a probability distribution over actions and/or function $Q_\phi(a|s)$ that defines what $R_n$ the agent is expecting to receive if they take action $a$.
Parameters $\phi$ are learned empirically, using gradient descent or evolutionary methods \cite{deeprl-survey1,deeprl-survey2}.

This approach has been applied extensively and with great success \cite{rl-survey} in Partially Observable Markov Decision Process (POMDP) settings, however it does have major limitations:
\begin{enumerate}
    \item The agent is defined as stateless. As such, when making a decision $a_i$ the agent is unable to take into account any observations it made prior to step $i$. Long-term dependencies like "this patient should not receive this drug since she has shown signs of allergy when this drug was administered to her 17 iterations ago" cannot be captured by a memoryless model.
    \item The agent is represented as a set of model weights $\phi$, often with millions of parameters. Such a program can be used as a black box decision system, but domain experts are unable to understand and/or make their contributions to the agent's programming.
\end{enumerate}

In this paper, we address these limitations by representing an RL agent with a program in a specialized language, to be introduced in section \ref{sec:bf}, as opposed to $\pi_\phi$ and $Q_\phi$

%% file: 3-relatedwork.tex
\section{Related work}
\label{sec:review}

Despite Program Synthesis being one of the most challenging tasks in Computer Science, many solutions exist, see \cite{synthesis-review}.
They can be roughly classified by specification modality: how are the requirements for a program to be synthesized communicated to the generative model?

The most advanced program synthesis technology to date is deep neural network-based language modeling \cite{codex}. 
This models are autocomplete engines that given a fragment of a program, also known as a prompt, predict the fragment to follow afterwards.
Such models can be very powerful, however, the prompt is a suboptimal form of specification, creating an open problem of prompt engineering - generating the first fragment of the program in such a way that it encourages the language model to write a particular prompt \cite{unreasonable}.

If the requirements are specified as a natural language description of what a program should do, program synthesis becomes a machine translation task \cite{sqlnet}.
A neural model can be pre-trained as a language model and fine-tuned on a translation dataset like CoNaLa \cite{conala} or, in the case of AlphaCode \cite{alphacode}, a dataset of competitive programming tasks and submitted solutions.

If the requirements are specified as a set of inputs to the program along with expected outputs, the task is known as programming by example \cite{nps-review,flashmeta} using techniques like neural-guided program search \cite{neural-guided}.
One can also generate input-output pairs artificially \cite{io-synthesis}.
Models like Neural Turing Machines \cite{neural-turing}, Memory Networks \cite{memorynets} and Neural Random Access Machines \cite{neural-ram} are also trained with input-output pairs and even though they don't explicitly generate code, they fit the definition of program.

In this work our goal is to synthesize programs with no explicit specification - only an environment where the program can be tested. 
This task is typically tackled with neural program synthesis \cite{pirl,brain-coder} or genetic programming \cite{genprog-applications} in a domain-specific language, where the reward function of the POMDP is known as the \emph{fitness function}.
However, to the best of our knowledge, there is no programming language for POMDP settings specifically.
Because of this, applications of genetic programming for Reinforcement Learning challenges have been limited and \cite{brain-coder}, for example, only supports non-interactive programs, i.e. a program is a mapping from an input string to an output string.

%% file: 4-language.tex
\section{BF++}
\label{sec:language}

\subsection{BF syntax}
\label{sec:bf}

Abolafia {\sl et al} \cite{brain-coder} picked BF\footnote{Brainfuck} \cite{brainfuck} as their language for program synthesis for the following reasons:
\begin{itemize}
    \item In industry-grade programming languages like Python or Java program code can contain a very large variety of characters since any of the 143859 Unicode \cite{unicode} characters can be used in string literals. In BF, however, only 8 characters can be used: they can be one-hot-encoded with vectors of size 8. 
    \item BF's simple syntax means that an arbitrary string of valid characters is likely to be a valid program. 
    In more complex languages, most possible strings result in a syntax error. 
    A generative model being trained to write programs in such a language risks being stuck in a long exploration phase when all the programs it generates are invalid and it has no positive examples in the dataset.
    \item Despite all of the above, it is a Turing-complete language.
\end{itemize}

The simplicity of the language also means that it is relatively easy to develop a compiler that translates programs from an industry-standard programming languages like Java and Python to BF thus making use of the expert knowledge existing in those languages. 

In the current paper, we introduce an extended version of the original BF language, BF++. As explained below, the extensions to the original BF syntax are particularly useful in the RL use cases. 

BF's runtime model is inspired by the classic Turing Machine \cite{turing}: at any point during the program's execution, the state of the program consists of:

\begin{itemize}
    \item An infinite\footnote{If you happen to be executing a BF program on a computer with finite memory, the tape will be finite due to your hardware limitations.} tape of cells $T$ where each cell holds an integer number.
    \item A \textit{memory pointer} $p_T$ that points to a certain cell in the tape (\textit{active cell} $T^{p_T}$).
    \item A string of characters $C$ that represents program code.
    \item A \textit{code pointer} $p_C$ pointing to a character about to be executed.
\end{itemize}

The code pointer starts at the first character, then this character gets executed and the pointer is incremented (moved to the next character).
There are 8 possible characters:

\begin{description}
\item[\texttt{>}] Move the memory pointer one cell right. $p_T := p_T + 1$
\item[\texttt{<}] Move the memory pointer one cell left. $p_T := p_T - 1$
\item[\texttt{+}] Increment the \textit{active cell}. $T^{p_T} := T^{p_T} + 1$
\item[\texttt{-}] Decrement the \textit{active cell}. $T^{p_T} := T^{p_T} - 1$
\item[\texttt{.}] Write $T^{p_T}$ from the \textit{active cell} to the \textit{output stream}\footnote{The definition of input and output streams is purposefully underspecified, it may depend on the particular implementation.}
\item[\texttt{,}] Read $x$ from the \textit{input stream} to the \textit{active cell}. $T^{p_T} := x$
\item[ \texttt{[} ] If the \textit{active cell} $T^{p_T} = 0$, jump (move $p_C$) to the matching $]$.
\item[ \texttt{]} ] If the \textit{active cell} $T^{p_T} \neq 0$, jump (move $p_C$) to the matching $[$
\end{description}

[ and ] commands constitute a loop that will be executed repeatedly until the \textit{active cell} becomes zero.
They are also the only way to write a BF program with a syntax error: a valid BF program is one that doesn't contain non-matching [ or ]


\subsection{Negative values}

In \textbf{BF} memory cells $T^i$ hold non-negative values only.
In \textbf{BF++} $T^i \in \mathbb{Z}$, a negation operator \texttt{\~} is introduced and operators \texttt{[]}are redefined to loop while the \textit{active cell} is non-positive, i.e.

\begin{description}
\item[ \texttt{\~} ] If the \textit{active cell} $T^{p_T} := - T^{p_T}$.
\item[ \texttt{[} ] If the \textit{active cell} $T^{p_T} \geq 0$, jump (move $p_C$) to the matching $]$.
\item[ \texttt{]} ] If the \textit{active cell} $T^{p_T} < 0$, jump (move $p_C$) to the matching $[$
\end{description}

This decision was taken because negative observations are common in control problems (see section \ref{sec:experiments}) as is branching on whether the observed value is positive or negative. 

\subsection{Non-blocking action operators}
\label{sec:queue}


The main issue of \textbf{BF} as a language for Reinforcement Learning is its input-output system.
It assumes that the program can freely decide on the relative frequency of inputs to outputs.
For example, the following program

\begin{center}
\begin{BVerbatim}
+[.....,]
\end{BVerbatim}
\end{center}

inputs 5 integers, outputs the 5th character it read, then goes back to the beginning and proceeds indefinitely outputting every 5th character it inputs.
Thus it assumes a 5:1 frequency of inputs to outputs.
If we simply assume that inputs are observations and outputs are actions, such program will not be able to operate in a POMDP environment where I/O frequency is fixed at 1:1 and the agent that has made an observation has to act before it can make the next observation.
In other words, operators \texttt{.} and \texttt{,} are blocking: \texttt{.} stops program execution and waits until new input is received to resume execution, \texttt{,} stops program execution and waits until there is an opportunity to act in the environment.

To address this, in \textbf{BF++} \texttt{.} operator is non-blocking.
It outputs the current value of the active cell by placing it at the bottom of the \textit{action queue} $S$ - a sequence of integer numbers that represent actions the program is planning to take in the environment. We also introduce a non-blocking operator \texttt{!} that places $T^{p_T}$ on top of the action queue.

\begin{equation}
    \begin{array}{cc}
         . & S := S^\frown (T^{p_T}) \\
         ! & S := (T^{p_T})^\frown S
    \end{array}
\end{equation}

where $\frown$ denotes concatenation of tuples

The program can thus decide by using \texttt{.} or \texttt{!} whether the newly added action takes precedence over ones already in the queue.
As soon as an opportunity to act arises, the top of the action queue (item $S^1$ or several items $S^1,S^2,\dots$, see section \ref{sec:envs}) defines which action the program takes and is then removed from the queue. 
If $S^k$ does not exist (the queue is empty or shorter than $k$) default value of $S^k=0$ is assumed.

\texttt{,} operator, on the other hand, is blocking. 
Thus its function is more important than just reading an observation into memory.
Executing \texttt{,} is when the program moves to the next step of POMDP.

\subsection{Virtual comma}
\label{sec:virtualcomma}


The system where the only way to proceed to the following iteration is the \texttt{,} operator, naively implemented, means that to be successful in any POMDP environment, a program has to contain an infinite loop with a \texttt{,} operator.
Any program that has a finite number of \texttt{,} steps will terminate prematurely in an environment that supports arbitrarily long number of iterations.
Since we originally set out to develop a language where most random programs would be valid, this had to be addressed.

We decided to turn any \textbf{BF++} program into an infinite loop with a \texttt{,} operator by default:
\begin{enumerate}
    \item Every \textbf{BF++} program starts with a virtual \texttt{,} operator at address $p_C = -1$: it is executed before all operators in the code of the program, they are indexed starting from $p_C = 0$
    \item When the code pointer $p_C$ reaches the end of the program it loops back to the virtual comma $p_C := -1$
\end{enumerate}

Due to the virtual comma, every program starts executing with the initial observation already stored in memory and available for branching/decision-making.

\subsection{Observation discretization}
\label{sec:observe}

Another issue complicating applications of \textbf{BF} to Reinforcement Learning is that since its memory tape holds only integer numbers its inputs and outputs have to be integer as well.
And this issue cannot be fixed simply by replacing an integer tape with a tape of floating point numbers as \textbf{BF}'s only operations for manipulating numbers are \texttt{+} and \texttt{-} - increment and decrement.
Non-integer action and observation spaces are fairly common in reinforcement learning tasks hence \textbf{BF++} implements coercion mechanisms for reading and writing continuous vectors into discrete memory.

We assume that the vector observation space $O$ is a hypercube defined as an intersection of $n$ separate scalar observation spaces $O^k$ such that 
\begin{equation}
 o_1 \in O_1^k,o_2 \in O_2^k,\dots,o_n \in O_n^k \Leftrightarrow (o_1,o_2,\dots,o_n) \in O  
\end{equation}

This assumption theoretically excludes some possible observation spaces, but almost all POMDP tasks discussed in the research literature and all OpenAI Gym tasks conform to this assumption.

To write an observation onto the memory tape we the observation vector of size $n$ is aligned with memory cells $T^{p_T},T^{p_T+1},\dots,T^{p_T+n-1}$ and turned into an integer with the use of $d$ discretization bins.

\begin{equation}
\label{eq:discretization}
T^{p_T+k-1} := \min_{\omega \in 1,\dots d | o^k < \tau^k_\omega} \omega
\end{equation}

If $O^k$ is an interval $O^k=[o_{low}, o_{high}]$, it is split into discretization bins evenly, as in eq. \ref{eq:static-thresholds}:

\begin{equation}
\label{eq:static-thresholds}
\tau_\omega = \begin{cases}
o_{low}+\frac{o_{high}-o_{low}}{d}\omega, \omega=1,2,\dots,d-1 \\
+\infty, \omega=d 
\end{cases}
\end{equation}

\begin{figure}[htb]
    \centering
    \includegraphics[width=0.8\linewidth]{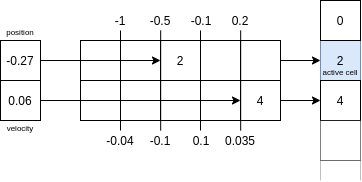}
    \caption{Fluid discretization example in Mountain Car}
    \label{fig:obs}
\end{figure}

Some environments, however, have unbounded observation spaces $O^k=(-\infty;+\infty)$, $O^k=(-\infty;o_{high}]$,  $O^k=[o_{low};+\infty)$.
This spaces are challenging because the formal description $O^k$ does not in any way reflect the actual underlying distributions of observations.
It can be the case, for example, that $O^k=(-\infty;+\infty)$ but most observations found in the environment fall in the interval $O^k=[42;43]$.
For such observation spaces, \textbf{BF++} uses a \textit{fluid discretization} system that learns the true distribution of observations online. The idea was inspired by a work of Touati {\sl et al} \cite{adaptivediscretization}, although, they assumed that $O^k$ has a finite diameter and didn't support unbounded observation spaces.
Initial thresholds $\tau_\omega$ can be arbitrary.
With each new observation, thresholds $\tau_\omega$ are readjusted so that among $h$ prior observations, roughly $\omega$ out of $d$  observations are lower values that $\tau_\omega$:

\begin{equation}
\underset{\tau}{\text{minimize}} \sum_{\omega \in 0,1,\dots d} |\frac{\omega}{d} - \frac{\sum_{i' \in i-h,i-h+1,\dots,i-1} \mathbb{I}(o_{i'}^k < \tau_\omega)}{h}|
\end{equation}

To solve this optimization problem, one has to sort previous $d$ observations in ascending order so that 

\begin{equation}
    \text{sort}: \{o_i | i \in i-h,i-h+1,\dots,i-1\} \longrightarrow \{ s_i | i \in 1,2,\dots,h \}
\end{equation}

is such a bijection that $s_1 < s_2 < \dots < s_h$ holds and set

\begin{equation}
    \tau_{\omega} = s_{\lceil \frac{\omega}{d} h \rceil}
\end{equation}

See figure \ref{fig:obs} for a visual example.


This system has 2 hyperparameters: $d$ and $h$.
With a low $d$ a lot of the information observed form the environment is lost, while when $d$ is in the hunderds the generated programs can become very complex.
$h$ switches between relative and absolute observations.
With a very high $h$, $\omega=0$ means that this observation is one of the lowest that can be observed in this environment, with $h=1$ it means that the observation is lower than the previous one.

High values of $h$ present an additional challenge: how to correctly discretize observation in the first $h$ iterations?
We implemented \textit{burn-in}: before training or evaluation we run $h$ iterations of a random agent (see section \ref{sec:random}) to collect a history of $h$ observations and pick correct thresholds.

\subsection{Action coercion}
\label{sec:act}

A symmetrical problem arises with actions taken by the agent. 
Memory tape holds integer numbers $T^k \in \mathbb{Z}$ and any value can be pushed onto the action stack.
However, the action that's output to the environment has to belong to a $N$-dimensional action space $A$, an intersection of unidimensional action spaces $A^k$.
The "act" operation thus includes a coercion system and is defined as:

\begin{equation}
\label{eq:act}
\begin{array}{l}
    a^k := \begin{cases}
\frac{S^k}{d-1}, A^k = (- \infty; + \infty) \\
a_{\text{min}} + |\frac{S^k}{d-1} - a_{\text{min}}|, A^k = [a_{\text{min}}; + \infty) \\
a_{\text{max}} - |a_{\text{max}} - \frac{S^k}{d-1}|, A^k = (- \infty; a_{\text{max}}] \\
a_{\text{min}} + \frac{(S^k \bmod d)}{d - 1} * (a_{\text{max}} - a_{\text{min}}), A^k = [a_{\text{min}}; a_{\text{max}}] \\
S^k, A^k \subset \mathbb{Z}
\end{cases} \\
    S := (S^{N+1}, S^{N+2}, \dots)
\end{array}
\end{equation}

\subsection{Goto}
\label{sec:goto}

It is notoriously hard to introduce any kind of branching behavior in \textbf{BF} \cite{brainfuck-control}.
To facilitate if-then style programs we introduce a \textit{goto} operator \verb|^| defined as 

\begin{equation}
   p_T := T^{p_T} 
\end{equation}

Note that it is not a \texttt{goto;} in the traditional C sense, since the memory pointer is being moved, not the code pointer.
Still, it lets the agent preemptively store potential actions in memory cells and than branch between this actions based on the observation.

\subsection{Random number generator}
\label{sec:random}

Operator \texttt{@} writes a random number into the \textit{active cell}.
A random agent is often used as a starting point for exploration and in \textbf{BF++} a random agent can be implemented as \verb|@!|

\subsection{Shorthands}
\label{sec:shorthands}

With all the commands we introduced in sections \ref{sec:bf} - \ref{sec:goto} it is still surprisingly hard to encode relatively simple decisions like "add action 5 to the top of the action queue":

\begin{center}
\begin{BVerbatim}
[>]+++++!
\end{BVerbatim}
\end{center}

This program moves the memory pointer right until it hits a cell that contains zero, increments it five times, and then pushes $T^{p_T}$ to the top of the action queue. It also loses the current value of the memory pointer which might be meaningful. Our experiments have shown that it takes a very long time for the neural model to learn to write this kind of combinations.

To mitigate this issue we introduce \textit{shorthands}: commands \texttt{01234} mean "write the respective number (0,1,2,3 or 4)" into the \textit{cell} and commands \texttt{abcde} mean "move the memory pointer to cell a,b,c,d or e" where cells a,b,c,d and e are the first 5 cells in the memory tape.
We intentionally made the number of \textit{shorthands} equal to discretization constant $d=5$.
Due to our method of discretization of continious action spaces (see sections \ref{sec:observe}, \ref{sec:act}) the program will often encounter situations when it can choose between $d$ different actions and thanks to shorthands taking them can be encoded as \texttt{1!}, \texttt{2!}, \dots

\subsection{Summary}
\label{sec:summary}

In total (assuming 5 shorthands) \textbf{BF++} has 22 commands:

\begin{center}
\begin{BVerbatim}
><^@+~-[].,!01234abcde
\end{BVerbatim}
\end{center}

Commands \verb|@^~01234abcde| are considered optional and can be disabled if the task at hand calls for it.
The number of shorthand commands can be increased or decreased.

Observation discretization and action coercion techniques built into the language mean that \textbf{BF++} is compatible with any POMDP environment. 
However, in practice, there is one important limitation: the complexity of the program required to operate in an environment is directly proportional to dimensionality of it's action and observation spaces $A$ and $O$. 
If, for example the observation space is 10000-dimensional, once an observation is read onto tape $T$ it takes 9999 \verb|>| operators to reach second to last observation.
Thus, in practice, \textbf{BF++} should be used with low-dimensional POMDPs.

An extension of our methodology to high-dimensional POMDPs (such as Atari games \cite{atari}, where the observation is a matrix of pixels on simulated game screen) can be achieved by adding a scene encoder neural network that maps the observed image to a low-dimensional vector as proposed in \cite{daqn}.

%% file: 5-experiments.tex
\section{Experimental setup}
\label{sec:experiments}

\subsection{Hypotheses and goals}
\label{sec:exgoals}

Our experiments were designed to test the following hypotheses:

\begin{description}
    \item[$H_1$] \textbf{BF++} can be used in conjunction with a program synthesis algorithm to solve arbitrary reinforcement learning challenges (POMDPs)
    \item[$H_2$] \textbf{BF++} can be used to take a program written by an expert and use program synthesis to automatically improve it
    \item[$H_3$] \textbf{BF++} can be used to generate an interpretable solutions to Reinforcement Learning Challenges that experts can learn from
    \item[$H_4$] Optional commands \verb|@^~01234abcde| introduced for convenience make it easier for experts to write programs in \textbf{BF++}
    \item[$H_5$] Optional commands \verb|@^~01234abcde| improve the quality of programs synthesised by neural models
\end{description}

Hence we

\begin{enumerate}
    \item Pick several commonly studied reinforcement learning environments
    \item Employ an expert\footnote{first author of this paper} to write \textbf{BF++} programs to solve them
    \item Develop a program synthesis model following from \cite{brain-coder}
    \item Compare the best programs generated by the model with expert programs in terms of program quality
    \item Perform ablation studies: remove some of the optional commands from the language (resulting language is called \textbf{BF+}), remove the expert program from the model's program pool, compare program quality
    \item Perform case studies: analyze programs generated by the model to gain insight into how the model approached the problem
\end{enumerate}

\subsection{Environments}
\label{sec:envs}

\begin{figure*}[t]
    \centering
    \subfloat[CartPole-v1]{
        \centering
        \includegraphics[height=1.3in]{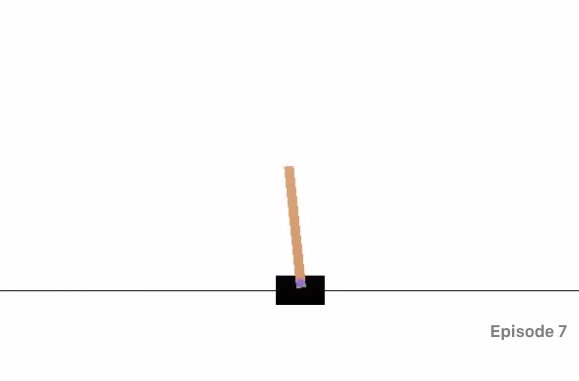}
    }
    \subfloat[MountainCarContinuous-v0]{
        \centering
        \includegraphics[height=1.3in]{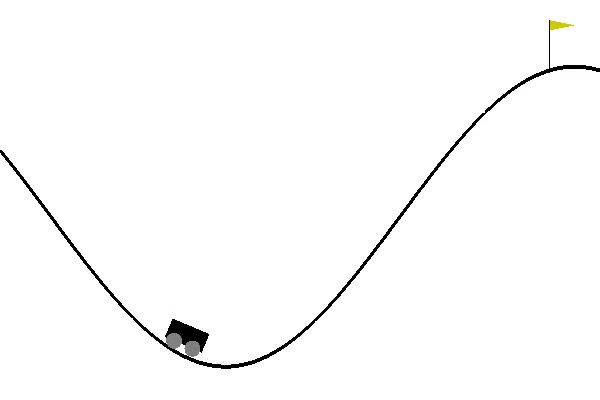}
    }
    \subfloat[Taxi-v3]{
        \centering
        \includegraphics[height=1.3in]{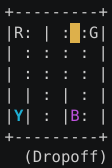}
    }
    \subfloat[BipedalWalker-v2]{
        \centering
        \includegraphics[height=1.3in]{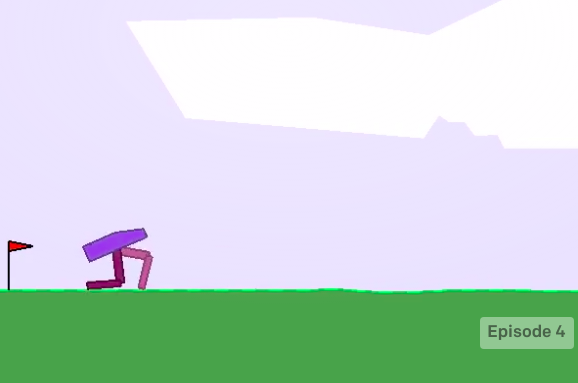}
    }
    \caption{Selected environments, visualized}
    \label{fig:envs}
\end{figure*}

We evaluate our framework on 4 low-dimensional (see section \ref{sec:summary}) POMDPs sampled from OpenAI Gym \cite{openai-gym} leaderboard\footnote{\url{https://github.com/openai/gym/wiki/Leaderboard}}:

\begin{enumerate}
\item \textbf{CartPole-v1} \cite{cartpole}.
A pole is attached to a cart.  which moves along a frictionless track.
The agent observes cart position, cart velocity, pole angle and pole velocity at tip.
The goal is to keep the pole upright by applying force between -1 and 1 to the cart.
At every step the agent receives a +1 reward for survival.
The episode terminates when the pole inclines too far.
\item \textbf{MountainCarContinuous-v0} \cite{mountain_car}.
A car is on a one-dimensional track, positioned between two "mountains". 
The goal is to drive up the mountain consuming a minimal amount of fuel by controlling the engine, setting it's torque in the range $[-1;1]$; however, the engine is not strong enough to scale the mountain in a single pass.
Therefore, the only way to succeed is to drive back and forth to build up momentum. 
We picked MountainCarContinuous-v0 as opposed to MountainCar-v0 to demonstrate the performance of our discretization system.
\item \textbf{Taxi-v3} \cite{taxi}. There are 4 locations (labeled by different letters) and the goal is to pick up the passenger at one location and drop him off in another in as few timesteps as possible spending as little fuel as possible.
\item \textbf{BipedalWalker-v2}. A simulated 2D robot with legs has to learn how to walk. 
Moving rightwards is rewarded, falling is penalized.
Observation vector consists of speeds, angular speeds and joint positions collected by the robot's sensors.
These observations do not, however, include any global coordinates - they can only be inferred from sensor inputs.
With action vector of size 4 the agent controls speeds of the robots hip and knee motors.
\end{enumerate}

\subsection{Hyperparameters}

For observation discretization (section \ref{sec:observe}) we picked $d=5$ (so that it's equal to the number of shorthands) and $h=500$ for our experiments, hence when the observation is among the highest 20\% of the last 500 observations it is written into memory as 4 while if it falls between 40-th and 60-th percentiles it is 2.

\subsection{Expert programs}
\label{sec:expert-progs}

For \textbf{CartPole} we wrote 2 programs. 
One completely ignores all observations and just alternates between "move right" and "move left":

\begin{center}
\begin{BVerbatim}
0!,1!
\end{BVerbatim}
\end{center}

Another calculates the difference between velocity of the cart and angular velocity of the pole.
If it's positive, the cart is pushed to the right (the cart has to catch up with the pole), if it's negative the cart is pushed to the left, if zero it is pushed randomly:

\begin{center}
\begin{BVerbatim}
[a0>0>0>0>0>@>1>1>1>1>1>,>[->>-<<]>>+++++^!1]
\end{BVerbatim}
\end{center}

The first part of this program sets up an action map on the tape where every possible value of the velocity differential has a respective cell with 0, 1 or (in the center) random number.
Then \verb|[->>-<<]| block does subtraction, \verb|+++++| adds 5 to the result, so that it belongs to in $0..10$ and not $-5..5$, \verb|^| moves the memory pointer to the correct cell in the action map and \verb|!| puts the action onto the action stack.

For \textbf{Mountain Car} we wrote an elegant algorithm that reads the observation vector into the tape, goes to the second observation (car velocity) and outputs it as action:

\begin{center}
\begin{BVerbatim}
>!a
\end{BVerbatim}
\end{center}

In other words, we apply motor torque in the same direction where we're currently headed, thus always accelerating our car.
If we're headed right, that helps us get to the destination and if we're headed left that helps us get as high as possible onto the hill so that when direction reverses, the car has more energy to push through the right hill.

For \textbf{Taxi} we introduce 2 programs.
The first program:
\begin{enumerate}
    \item Finds the coordinates of the current destination (passenger to pick up or current passenger's destination)
    \item Subtracts the current destination 
    \item Moves in the resulting direction
\end{enumerate}

The problem with this approach is that it always gets stuck when it hits a wall.
To compensate for that, the second program alternates between the strategy above (for 5 iterations) and random movements (for 5 iterations) so that it eventually gets unstuck. See source code repository for the programs.

Optional commands \verb|@^~01234abcde| have all been invaluable in developing these programs - a fact in support of $H_4$.
A more rigorous way to confirm it would be employing several human experts to develop programs with and without optional operators, but finding volunteer \textbf{BF++} developers has proven difficult. 

Developing programs for \textbf{Bipedal Walker} is, unfortunately, above our expert's paygrade. 

\subsection{Program synthesis model}
\label{sec:methodology}

In order to train a generative model $g$ to write \textbf{BF++} programs we treat the writing process as a reinforcement learning episode in its own right \cite{brain-coder} .
Every character of a program is an action taken by the \emph{writer agent}, the programs are terminated by a NULL character.
When the NULL character is written, a \emph{BF++ agent} is created in the target POMDP environment (e.g. CartPole) and sum total of rewards $Q$ collected in that episode is assigned as a reward to the \emph{writer agent} for the NULL character.
All other characters are rewarded with zero.

The \emph{writer agent's} policy is modeled with an LSTM \cite{lstm} neural network and is trained with a modified version of REINFORCE \cite{reinforce} algorithm.
While standard REINFORCE optimizes Policy Gradient:

\begin{equation}
    O_{\text{PG}}(\phi) = \mathbb{E_{\pi(C; \phi)}}(Q)
\end{equation}

where $\phi$ are LSTM parameters, $C$ - program, $Q$ - reward obtained by the program in target environment,

we optimize

\begin{equation}
    O(\phi)=O_{PG}(\phi)+O_{PQT}(\phi)
\end{equation}

where

\begin{equation}
    O_{\text{PQT}} = \frac{1}{K} \sum_{k=0}^K \log \pi(C_k; \phi)
\end{equation}

where $C_1$ is the best (highest $Q$) known program, $C_2$ - second best, \dots

Intuitively, both $O_{\text{PG}}(\phi)$ and $O_{\text{PQT}}(\phi)$ when optimized update the weights of the LSTM so that programs that we have found to be successful are more likely.
But Policy Gradient weighs programs proportionately to their respective rewards while PQT creates a \textit{priority queue} of the \textit{best known programs} and assigns a high importance to them and zero to the rest.

\begin{figure}
    \centering
    \includegraphics[width=\linewidth]{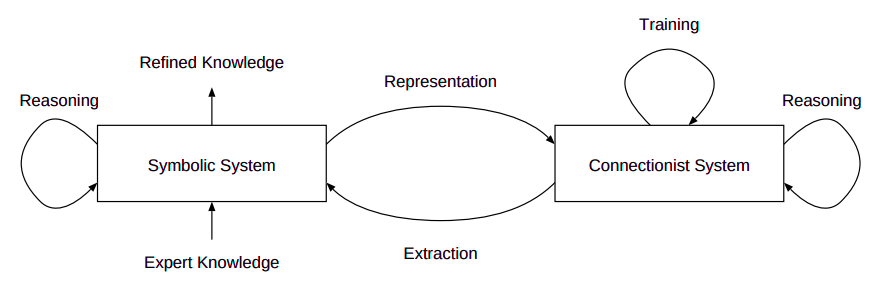}
    \caption{Neural-symbolic learning cycle \cite{cycle}}
    \label{fig:cycle}
\end{figure}

$O_{\text{PQT}}$ component has been shown to have "a stabilizing affect and helps reduce catastrophic forgetting in the policy" \cite{brain-coder}.
In addition to this, we use $O_{\text{PQT}}$ to implement \textbf{expert inspiration}.
By default, the \textit{priority queue} of the \textit{best known programs} is initialized as an empty set.
But if expert-written programs are available, it can be prepopulated with these programs that act as useful positive examples for teaching the \emph{writer agent}.
This approach is used to incorporate programs from section \ref{sec:expert-progs} and transfer knowledge from experts to the neural developer.

This approach to \textbf{expert inspiration} follows what's known as neural-symbolic learning cycle, displayed in figure \ref{fig:cycle} - expert knowledge is represented symbolically, in terms of a \textbf{BF++} program, then a neural network is trained to generate this program, effectively translating the expert knowledge from symbolic into connectionist format (\emph{representation}), the neural network learns from reinforcement how to solve the task better than the expert (\emph{training}).
Unlike in most neural-symbolic systems \cite{neuralsymbolic} that extract knowledge from connectionist systems with algortihms like TREPAN \cite{trepan} or JRip extraction \cite{jripextr}, the \emph{extraction} step is trivial since the neural network outputs a symbolic program directly.

In all experiments below, the \emph{writer agent}'s LSTM has hidden size of 50, batch size of 4 and is trained with RMSProp \cite{rmsprop} optimizer.

\subsection{Stopping and Scoring}

All experiments were run with an upper limit of 100000 training episodes.
Environments other than \textbf{Taxi} also used Exponential Variance Elimination \cite{evestop} early stopping technique - training was stopped when the postive trend in the quality of the best found program stopped, i.e. when the exponential moving average of program quality is lower that it was 1000 episodes ago.
Agents for \textbf{Taxi} are trained for a fixed number of episodes, because we noticed that in this environment the longest part of the training process is learning to pick up your first passenger and until that happens $Q=-200$ holds.

Once the training process is finished, we take the best known programs and since each of them was only tested once (leading to high variance) we test them again, averaging total rewards over 100 episodes. 
We use this averaged reward to pick the best program.

\subsection{Implementation}

\textbf{BF++} interpreter and the training system were written in Python with TensorFlow for neural models.
GPU resources weren't used, because the performance bottleneck of the system is not backpropagation but rather testing a \textbf{BF++} program in the environment, single experiment runtime was between 1 hour (CartPole) and 10 (Taxi).

%% file: 6-results.tex
\section{Results}

\begin{table*}[tb]
  \caption{Total episode reward $Q$ achieved by best programs found, averaged over 100 episodes}
  \label{tab:quality}
  \centering
  \begin{tabular}{lcccc}
    Environment     & CartPole-v1     & MountainCarContinuous-v0 & Taxi-v3 & BipedalWalker-v2 \\
    \midrule
    Random agent & 9.3 & 0 & -200 & -91.92  \\
    \midrule
    BF++ expert program 1 & 20.48 & -6.55 & -179.49 & - \\
    BF++ expert program 2 & 18.23 & - & -150.44 & - \\
    BF+ (without shorthands) LSTM & 44.55 & 91.57 & -57.93 & -91.9 \\
    BF+ (without \verb|@^~|) LSTM & 48.14 & 81.16 & -42.21 & -31.79 \\
    BF++ LSTM     & 71.38 & 88.41 & -199.82 & -26.97 \\
    BF++ LSTM with expert inspiration  & 96.64 & 91.39 & -60.65 & - \\
    \midrule
    Leaderboard threshold & 195 & 90 & 0 & 300 \\
    \bottomrule
  \end{tabular}
\end{table*}

\subsection{Quantitative results}

Table \ref{tab:quality} presents the quality metric (average 100-episode reward) of the best program in every category, compared to that of a fully random agent and the result required to join the OpenAI gym leaderboard for context.
Note that the expert programs used a lot of optional operators (shorthands and \verb|@^!|), so it wasn't possible to implement expert inspiration with limited command sets.

These results support (see section \ref{sec:exgoals}) hypothesis $H_1$ - we have obtained functional programs for all environments, $H_2$ - when expert inspiration was used the resulting programs were better than expert programs and better than programs generated without expert inspiration and $H_4$ - ablation studies for optional operators do indeed show that those operators are useful.

\subsection{Case studies}
\label{sec:casestudies}

We have established that the program synthesis model is able to learn from human experts.
But can experts learn from the model? ($H_3$) 
To confirm this, we offer a detailed explanation of the most successful program of all experiments listed in section \ref{sec:experiments}.

This program scored \emph{91.39} on \textbf{Mountain Car}:

\begin{center}
\begin{BVerbatim}
-..~+
\end{BVerbatim}
\end{center}

The trailing \verb|~| and \verb|+| do not affect the behavior of the agent: they modify the value of the active cell only for it to be immediately rewritten by the virtual comma (section \ref{sec:virtualcomma}) before it has any chance to influence actions.
One can think about these commands as inactive genes in the DNA - we have found many resulting programs to contain such commands.
If necessary this effect can be accounted for by incorporating program length into the loss function.
So this program is equivalent to:

\begin{center}
\begin{BVerbatim}
-..
\end{BVerbatim}
\end{center}

\begin{figure}
    \centering
    \includegraphics[width=\linewidth]{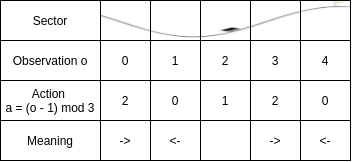}
    \caption{Visual summary of the strategy enacted by \texttt{-..} on \textbf{Mountain Car}}
    \label{fig:mountaincarwinner}
\end{figure}

When the virtual comma is executed, car position and car velocity are read into memory, discretized into integers $0\dots4$.
The position is read into the active memory cell $p_T$, while the velocity is in cell $p_T+1$.
Then the active cell is decremented and the resulting number is put onto the action stack twice.
There is 1 read operation and 2 write operations to the end of the action stack, which introduces a delay before the actions get executed.
When it's time to act, the number on the action stack is coerced to one of the actions possible in this environment (0 for going left, 1 for doing nothing, 2 for right). 

A strategy emerges, illustrated on figure \ref{fig:mountaincarwinner}, in which the car puts "going right" onto the agenda if it's on the far left or the center right of the landscape, puts "going left" onto the agenda when it's on the far right or center left and schedules doing nothing if it's in the center.
This strategy helps the car successfully reach the right fringe every time it is applied.

\section{Conclusions}

In this paper, we have introduced a new programming language tailored to the task of programmatically interpretable reinforcement learning.
We have shown experimentally that this language can facilitate program synthesis as well as knowledge transfer between expert-based systems and data-driven systems. 

The results in the OpenAI gym test examples show that the proposed system is able to find a functional solution to the problem. In some cases the performance is similar to the best deep learning solution but the obtained program remains still explainable. This is a very encouraging result and suggest that the use of program induction methods may indeed be a viable way towards explainable solutions in RL applications. 

We propose the following directions for future work:
\begin{enumerate}
    \item Develop translation mechanisms between \textbf{BF++} and other languages. Potentially, \textbf{BF++} can be used as \emph{bytecode} \cite{bytecode} for reinforcement learning. The expert would write a program in a higher-level language and transpile it into \textbf{BF++} so that the program then can be improved with reinforcement learning.
    \item Use other neural network architectures as well as non-neural evolution methods like genetic programming \cite{genprog} in conjunction with \textbf{BF++}
    \item Apply the framework to problems in Healthcare where expert inspiration is important for crossing the AI chasm \cite{aichasm}.
    \item Use Natural Language Generation techniques to translate the BF++ code automatically to a friendly human-readable text description as in \cite{code2nlg1,code2nlg2}.
\end{enumerate}